# Handwriting Recognition

[1]Jayati Ghosh Dastidar, [2]Surabhi Sarkar,
[3]Rick Punyadyuti Sinha, [4]Kasturi Basu

Department of Computer Science,
St. Xavier's College (Autonomous), Kolkata

*Abstract*—**This paper describes the method to recognize offline handwritten characters. A robust algorithm for handwriting segmentation is described here with the help of which individual characters can be segmented from a selected word from a paragraph of handwritten text image which is given as input.**

## I. INTRODUCTION

The issue of pattern recognition is central to many applications of computer science and technology. The off-line recognition of handwritten text is an interesting problem because it is easy to scan a handwritten document, to train the system with a particular handwriting style and let the system classify the rest of the text. In this paper the objective is to classify isolated words.

## II. OVERVIEW

The main idea is to extract letter features from handwritten letters in the corpus and to build a database directly from images. The features of new letters will be extracted and will be compared with those in the database.

The algorithm extracts the features directly from a bitmap or PNG image. Firstly, a line is extracted from the whole image and then each letter is extracted from that line. The letter is matched with the sample template stored in the database and the template with the highest correlation coefficient gives the output text. The steps we follow have been shown in the following diagram:

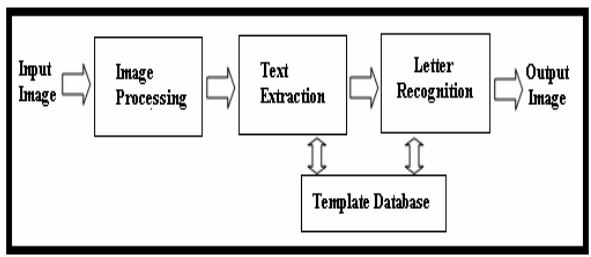

## III. DATA PREPROCESSING AND IMAGE EXTRACTION

Input images are taken from cursive handwritten text. Images are then filtered to obtain the best representation for each word. The document is scanned and a grayscale image is obtained for system input from the original RGB image. The grayscale image is then converted into a binary image after the computation of the threshold value. All objects having fewer than 15 pixels are removed then. This concludes the process of data preprocessing or noise removal.

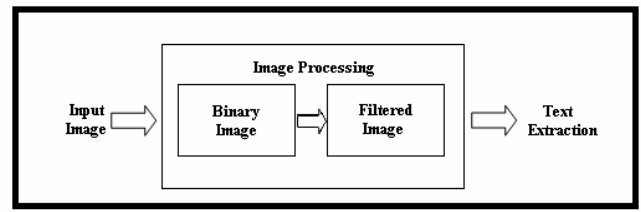

**Stages in the text recognition process**

## IV. LINE EXTRACTION

This initially considers the entire image after complete noise removal. The image is clipped so that the area of a negative binary image that has no white points in it is filtered out. The process of line extraction begins with storing the number of rows of the image matrix in a variable.

As the program works with a binary image, the two colors available in the entire image is either black or white. As the image is inverted the black points hold the value zero while the white points hold the higher value which is one. A check is made to see whether each row of the clipped image has any white points in it.

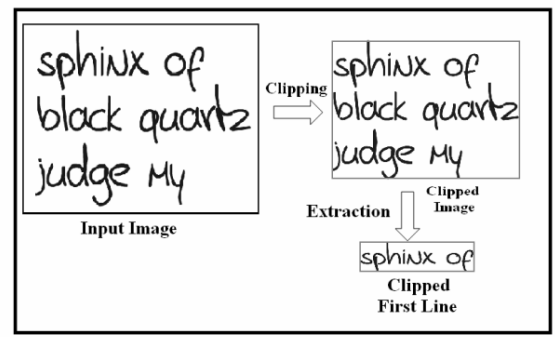

A blank line is detected when the sum of all the white points in the row is zero. The line above that row position is extracted out which becomes the first line and the remaining image again goes through the same process to extract the next set of lines.

The first line obtained is clipped before performing letter extraction.

## V. LETTER EXTRACTION

This initially considers the entire image that is under consideration after the line extraction i.e., the first line. The process of letter extraction begins with storing the number of columns of the image matrix in a variable.

A check is made to see whether each row of the clipped image has any white points in it. A blank space is detected when the sum of all the white points in the column is zero.

Then the letter is cropped out from the sentence under consideration. The column where the last black point is obtained gives a length from the starting column (say length1) and it is now clipped. This clipped length is then subtracted from length1 and the space before the extracted letter is obtained. This whole process is repeated till we extract all the letters from the sentence under consideration.

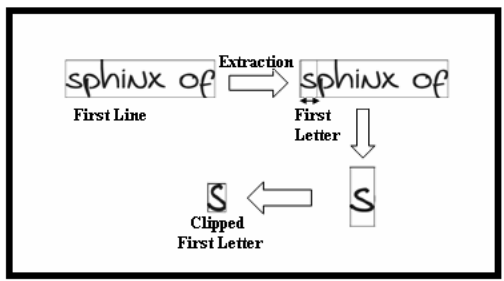

Thus the function begins to extract letters and stores the remaining portion of the line in another variable which is again a two-dimensional matrix. Both the first and the second matrix are clipped now to get the actual area completely covering the white points. The difference between the remaining portion of the line and the clipped remaining portion of the line is the space between any two letters. The space, first alphabet and the remaining portion of the line forms the output of this stage.

## VI. CLIPPING

The image is clipped to filter out the area of the image where no text is present. The clipped image is returned in the form of a two-dimensional matrix. The function first finds out the number of rows and columns present in the given image having a non-zero index. It then organizes the image and stores the value of the clipped matrix in another two dimensional variable matrix. The second variable matrix is then returned and hence the image gets clipped.

## VII. RECOGNITION

After extracting each letter from a line, an algorithm is needed to recognize the letters and convert them to text.

### A. Templates creation :

To achieve that, a database of all uppercase (A-Z) and lowercase (a-z) letters and numbers (0-9) are created as templates. The images are stored in a folder in .bmp or .png format with size <150 bytes and dimensions of 42X24 pixels. The images should be stored in the handwriting font to attain maximum correlation coefficient. Then a template store is created which is a cell matrix consisting of 62 cells with 42X24 dimensions.

### B. Reading the extracted letter:

It computes the correlation coefficient between the templates and the input image (extracted letter). The size of the input image should also be 42X24 pixels and if not then the algorithm resizes the image dimensions as per the above requirements.

The input image is compared with all the templates saved in the database. The template having the highest correlation coefficient with the input image becomes the output letter (A-Z or a-z) or number (0-9). This letter is then grouped with the upcoming recognized letters to form a word.

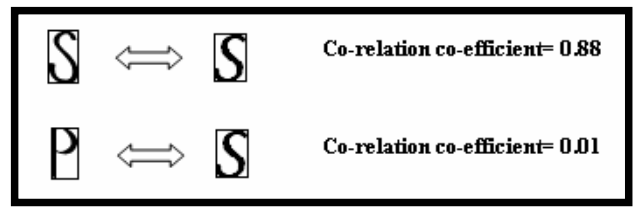

For example when an extracted 's' is compared with all the letters of stored in the database, the highest correlation coefficient is obtained when it is compared with the 's' stored in the database rather than any other letter like 'p' as shown in the above figure.

But some cases are exceptional which create a problem.

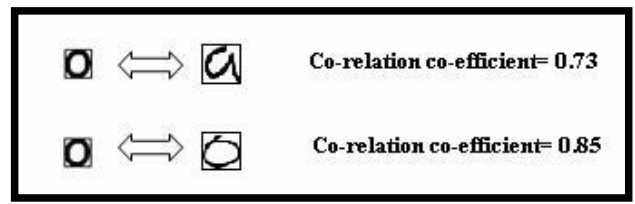

Like a handwritten 'a' has the highest correlation coefficient with the 'o' stored in the database just like the handwritten 'o'. So it recognizes the handwritten 'a' as 'o'. In these cases we need to extract more features from the handwriting.

### C. Assembling letters into words:

After the letters have been recognized, they will be brought together to form distinct words. The spaces obtained during the letter extraction process are introduced at the appropriate places to convert them into meaningful sentences.

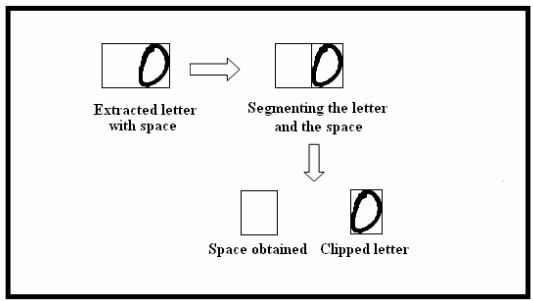

The space lengths obtained during letter extraction are stored in a 1-dimensional array. The spaces which are greater than or equal to 75% of the maximum space length are selected as a spacebar (' ') which is written in the output text file after a recognized word. This space obtained behaves as the delimiter between two words.

## VIII. Conclusion

This paper has highlighted a method for handwriting recognition which is promising and quite effective. It uses examples from cursive handwritten words as a bitmap image. It is clear that if we increase the number of database samples we increase the time taken by the algorithm linearly. The system is very general and hence we can also use this algorithm to recognize handwritten text in other languages by making modifications in the database.

Some shortcomings still persist in the algorithm and needs to be solved. Firstly, this algorithm will not work for connected cursive handwriting because the threshold value set for detecting the blank spaces between the letters is zero. The practical and optimal threshold value to check the density between the different connected alphabets for different handwriting styles can be obtained by further experimentation and observation of more sample handwritings. Secondly, since handwritten characters can be considered as a construction of line segments at different orientations and lengths. In future we hope to modify the feature vector with measurements obtained from filtered oriented parts.

## About the Authors

[1]**Jayati Ghosh Dastidar** (jghoshdastidar@gmail.com)

She has been a part of the education sector for over ten years. She is currently teaching in the Department of Computer Science, St. Xavier's College (Autonomous), Kolkata as an Assistant Professor. Her teaching experience includes both theoretical as well as practical Computer Science topics. She specializes in the teaching of courses such as Compiler Design, Automata Theory, Image Processing, Digital Electronics, Computer Organisation, Data Structure, etc.

[2]**Surabhi Sarkar** (0surabhi0@gmail.com),
[3]**Rick Punyadyuti Sinha** (ricksinha@rediffmail.com),
[4]**Kasturi Basu** (itsromi.kasturi@gmail.com)

They are third year students of the Department of Computer Science, St. Xavier's College (Autonomous), Kolkata. As students of the department, they have actively participated in a seminar on Data Mining and Warehousing in 2009 and a seminar on Modelling and Simulation Techniques for Computational Problems using MATLAB® in 2010.